\documentclass[10pt,twocolumn,letterpaper]{article}

\usepackage{wacv}
\usepackage{times}
\usepackage{epsfig}
\usepackage{graphicx}
\usepackage{amsmath}
\usepackage{amssymb}
\usepackage{multirow}
\usepackage{longtable}
\usepackage{supertabular}
\usepackage{soul, color}
\usepackage{booktabs}
\usepackage{multirow}
\usepackage{algorithm}
\usepackage[noend]{algpseudocode}
\usepackage{subcaption} 
\usepackage{verbatim}
\usepackage{float}
\usepackage{bm}
\wacvfinalcopy 

\def\wacvPaperID{****} 

\ifwacvfinal\fi
\begin{document}

\title{Spatio-Temporal Pyramid Graph Convolutions\\ for Human Action Recognition and Postural Assessment}
\author{Behnoosh Parsa  \\
University of Washington \\
{\tt\small behnoosh@uw.edu}
\thanks{Co-first authors who contributed equally to this work.
\newline This work was done while Behnoosh Parsa was an intern at Honda Research Institute USA.}
\and
Athma Narayanan \\
Honda Research Institute\\
{\tt\small anarayanan@honda-ri.com}\footnotemark[1]
\and
Behzad Dariush \\
Honda Research Institute\\
{\tt\small bdariush@honda-ri.com}
}

\maketitle
\ifwacvfinal\thispagestyle{empty}\fi

\begin{abstract}
Recognition of human actions and associated interactions with objects and the environment is an important problem in computer vision due to its potential applications in a variety of domains. The most versatile methods can generalize to various environments and deal with cluttered backgrounds, occlusions, and viewpoint variations. Among them, methods based on graph convolutional networks that extract features from the skeleton have demonstrated promising performance.   In this paper, we propose a novel Spatio-Temporal Pyramid Graph Convolutional Network (ST-PGN) for online action recognition for ergonomic risk assessment that enables the use of features from all levels of the skeleton feature hierarchy. The proposed algorithm outperforms state-of-art action recognition algorithms tested on two public benchmark datasets typically used for postural assessment (TUM and UW-IOM). We also introduce a pipeline to enhance postural assessment methods with online action recognition techniques. Finally, the proposed algorithm is integrated with a traditional ergonomic risk index (REBA) to demonstrate the potential value for assessment of musculoskeletal disorders in occupational safety.
\end{abstract}

\section{Introduction}
Human action recognition has been a widely studied research topic in computer vision for several decades.  The task is to infer the human action and activity from still images or video frames.  Solutions to this important and challenging problem have traditionally been applied to domains such as surveillance, entertainment, robotics, video retrieval, and intelligent driving assistance systems~\cite{poppe2010survey,oh2011large,zheng2019relational}. Recently, there are emerging applications that involve assessment of human performance for virtual fitness, health monitoring, training, and ergonomic risk assessment for occupational safety~\cite{parsa2019toward, guo2017fitcoach, prati2019sensors}. These applications have unique requirements that may involve simultaneous association of time varying pose with action and object interaction, and relating such information for computational modeling and prediction of various biomechanical indicators. Vision only systems are non-invasive and less expensive alternatives to study these problems as opposed to expensive drift prone motion capture systems and wearable sensors \cite{mazzoldi2002smart,cloete2008benchmarking}. 
\begin{figure}[t]
\begin{center}
\includegraphics[width=0.8\linewidth]{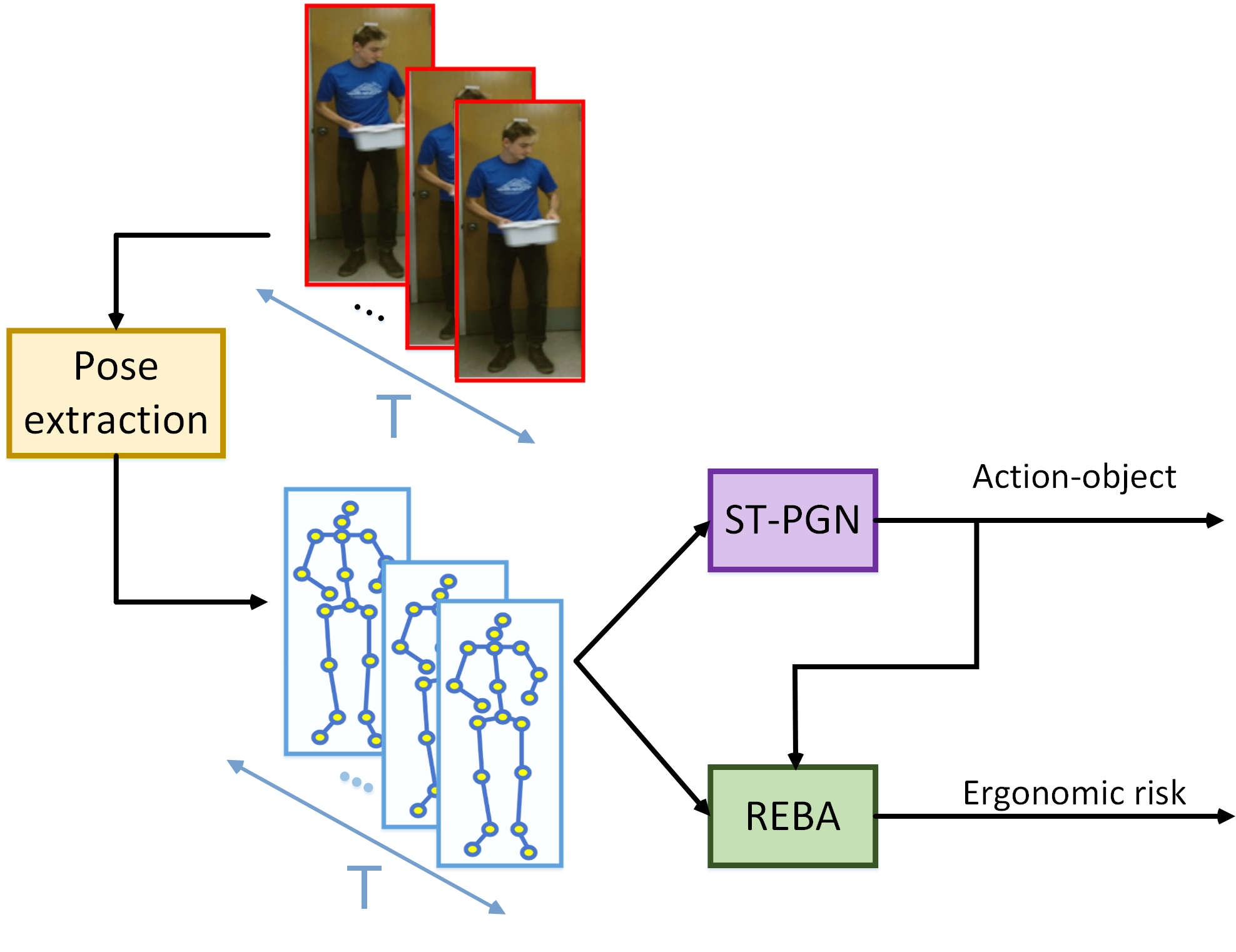}
\end{center}
  \caption{Our model (ST-PGN) takes a sequence of skeleton input produced by a pose extraction unit (like LCR-Net \cite{rogez2017lcr}) and does early action recognition. The skeleton sequence along with the activity labels go to the REBA computation unit to assess the ergonomic risk while testing.}
\label{fig:test_pipline}
\end{figure}

Depending on the application, human action recognition can be formulated in an online or off-line setting.  In most applications, processing is performed off-line, making use of the entire video sequence without strict limitations on computational resources.   In such cases, the typical assumption is that the start and end points of the action is known \cite{du2015hierarchical,mavroudi2018end} and the training video is pre-segmented into various action classes. Recent advances in hardware and GPU performance has led to the emergence of many online applications, where the requirement is to process video streams in real-time and without a priori knowledge of the transitions between actions  \cite{liu2018ssnet,liu2017global,singh2017online}.    
Generalization of action recognition algorithms is a challenging and unsolved problem.  Ideally, the method should generalize to various environments and deal with cluttered backgrounds, occlusions, and viewpoint variations. While end to end video to action classification have shown great promise, generalization is achieved through domain adaptation \cite{habibie2019wild,busto2018open} or using intermediate skeletal representations that are robust to these variations \cite{yan2018spatial,rogez2016mocap}. In particular, skeleton-based features appear to produce favorable results since human pose is typically a consistent representation of action across people and context.  Among them, recent work based on graph convolutional networks that extract meaningful features from the skeleton have achieved good performance~\cite{yan2018spatial,li2019actional}.   

The work in this paper is inspired by emerging applications involving human performance assessment in various domains including health, fitness, rehabilitation, and occupational safety.   In particular, we consider  specific challenges for real-time ergonomic risk assessment in complex environments such as manufacturing assembly.  The requirements include correlation of action with the time varying posture and associated ergonomic and biomechanical risk.   The ultimate goal is to produce reliable estimates of pose, action, and associated ergonomic indicators in order to identify the risk of musculoskeletal disorders associated with acute and repetitive tasks.   

To achieve this goal, we propose a novel real-time Spatio-Temporal Pyramid Graph Convolutional Network (ST-PGN) for action recognition that enables the use of  features from all levels of the skeleton feature hierarchy.  We tested the performance of the algorithm on two public benchmark datasets typically used for postural assessment (TUM and UW-IOM) as well as Kinetics and NTU-RGBD datasets.   The main contributions of this paper are as follows: First, we demonstrate the efficacy of the ST-PGN algorithm by achieving high recognition performance on long video sequences as well as the common public benchmarks.  We show that the algorithm is also able to learn the transitions between actions and is suitable for real-time applications.  Second, as compared to the state-of-the-art algorithms such as ST-GCN~\cite{yan2018spatial}, our model has fewer graph convolution kernels without sacrificing performance. Third, the feature pyramid architecture enables the proposed model to automatically capture the correlation between body parts, rather than hand-coding body-part relations.  Finally, we introduce a pipeline to enhance postural assessment methods with online action recognition techniques. The proposed ST-PGN algorithm is integrated with a traditional ergonomic risk index (REBA)~\cite{hignett2000rapid} to demonstrate the potential value for assessment of musculoskeletal disorders in occupational safety.

\section{Related Work}
Given the recent advances in obtaining accurate pose through depth sensing or vision based pose estimation algorithms, skeleton based action recognition methods have become vital for achieving generalization across a variety of environments~\cite{rogez2017lcr}.  Skeleton based methods also offer opportunity to study down stream applications that involve human performance analysis and require postural assessment.  We summarize work related to the proposed ST-GPN algorithm for association of action, posture, and ergonomic risk.  This section surveys the literature in \textit{action classification, graph convolution and ergonomic risk assessment.} To the best of our knowledge this is the first work that combines the three separately studied problems jointly and in an online fashion.
\vspace{-0.5em}
\subsection{Action classification} 
Video action segmentation tasks such as \cite{shou2016temporal,sultani2016if,zhao2018recognize,tran2018closer,carreira2017quo,tran2015learning,yan2018spatial} focus on localizing action labels in untrimmed videos or classifying entire video clips with one label. Evaluation is generally performed on large scale datasets such as  \cite{soomro2012ucf101,kuehne2011hmdb,idrees2017thumos,carreira2017quo}. Variations of temporal convolutions outperform conventional recurrent networks in these tasks \cite{bai2018empirical,lea2017temporal} since they are capable of aggregating  motion changes and long term temporal windows of past and future frames.  Extending these models to an ongoing, partially observed, and multi-action sequence, such as in online ergonomic risk assessment, is unclear. Hence, we focus on models and datasets (TUM \cite{tenorth2009tum} and UW-IOM \cite{parsa2019toward}) that translate to ergonomic risk assessment datasets that contain repetitive actions that are closely tied to activities such as manufacturing assembly. As evidenced by our ST-GCN\cite{yan2018spatial} experiments, these offline models do not translate well to other online ergonomic action datasets.  

Most similar to our work are \cite{liu2018ssnet,liu2017global,singh2017online}, which address online early action recognition for indoor datasets such as  \cite{li2016online,liu2017pku}. However, the focus of those works is on modelling the temporal evolution of poses and early prediction of future actions.  Rather than predicting future pose streams, our aim is to instead classify incoming pose streams. It is imperative to capture local label transitions (\textit{reaching} to \textit{pickup}) by exploiting subtle pose cues and temporal sequence understating. Hence, our focus is in designing a hierarchical architecture that can do these tasks jointly.  With advances in reliable pose estimation models \cite{zhao2017spindle,rogez2017lcr,cao2018openpose}, skeleton only action recognition has gained popularity \cite{zhang2017view,yan2018spatial,lee2017ensemble,si2019attention}.   Those methods have shown to be robust to variations in illumination and scene, and are typically context agnostic. 

One limitation of previous methods is that they do not contain the necessary features from scene context or object handling that give more meaning to the actions (e.g. walking on crosswalk means crossing verses walking indoors, lifting box vs lifting rod). Using scene only cues limits models from capturing complex pose dynamics and relative pose structure changes (e.g. hand moving in relation to torso means reaching for object). In this regard \cite{yan2018spatial} is, to our knowledge, the first method to operate on a local pose structure graph. 
However, our work addresses the sub-problem of online ergonomic-action classification by exploiting hierarchical spatio-temporal cues jointly. To focus of our discussion and avoid comparison to plethora of action recognition work, we compare our models to spatio-temporal models that use GCN.

\subsection{Graph Convolution Networks}
Graph convolution network (GCN) is a powerful method for processing non-Euclidean spaces \cite{wu2019comprehensive}.  Since the skeleton structure is inherently represented as a graph with nodes and connections, GCN is increasingly being used for analyzing human motion for different applications.
Spatio-temporal graph convolutions add another dimension to GCN by applying convolutions over spatial domain, and temporal convolutions (TCN) over the time domain in a sequential manner. Most related work in skeleton based action recognition include \cite{yan2018spatial,kim2019skeleton,li2019actional,si2019attention}. The first three papers focus on graph convolution on temporal skeleton sequences. However, they do not model the hierarchical parts structure in graphs.

Recently, Kim et al.\cite{kim2019skeleton} introduced a two-stream method for human action recognition. They used a human pose stream based on ST-GCN and an object-related pose stream which is achieved by training an object detector on the set of objects of their interest. Similarly, our work attempt to fuse the object/context features along with pose dynamics. However, we focus on enhancing the skeleton features and treat objects as features from VGG16. We propose an alternative strategy for fusion inspired by GRU. The focus is to avoid confusion between objects handled in the labels (pose configuration for \textit{rod-pickup} and \textit{box-pickup} look similar).

\subsection{Ergonomic risk assessment}
Work-related musculoskeletal disorders (MSDs) are costly, affect all age groups, and are common in many occupations. MSD is a major contributing factor to disability, loss of independence, and early retirement. Therefore, many studies analyze the ergonomic risk to workers, particularly in   manufacturing assembly \cite{peppoloni2016novel, singh2017ergonomic, colim2019ergonomic, roodbandi2018prevalence, possebom2018comparison,malaise2019activity}. 

Rapid Entire Body Assessment (REBA) \cite{hignett2004rapid} and The European Assembly Worksheet (EAWS) \cite{schaub2013european} are two common ergonomic risk measures used in the industry.  REBA assigns human posture scores, in the range 1-15, based on joint angles during an activity. First, a risk score is computed for lower and upper extremities and those scores are added to the task-related scores such as coupling and load. EAWS, however, is focused at the activity level, and how it is performed. Both metrics are traditionally determined visually, by an expert observing the action. 

Li et al. \cite{li2011computer} use distributed surveillance cameras and body-mounted motion sensors to automatically calculate ergonomic risk. Shafti et al. \cite{Shafti2019} use an RGB-D camera to understand the safe range for arm motions and give feedback on the subjects' performance during welding. Kim et al. \cite{Kim2019} use a camera to monitor and adjust the ergonomic risks of working with power tools in real-time. Parsa et al. \cite{parsa2019toward}, introduced an offline method to segment a video into semantically meaningful actions and report an ergonomic risk level for each action. Online ergonomic risk computation provides a real-time assessment based on the individual's posture.
Improved ergonomic assessment, particularly for metrics such as EAWS, can be attained by considering not only the posture, but also the action and object interaction.  In this work, we compute REBA frame-wise and use the recognition predictions to adjust the scores. Our activity recognition predicts the postures and actions, and identifies object interactions and the height at which the activity is being performed.  Such information affect the REBA score computation.

\section{Spatio-Temporal Feature Pyramid Graph Convolution}
In this work we introduce Saptio-Temporal Pyramid Graph Convolutional Network (ST-PGN). ST-PGN models the spatio temporal features of the skeletal structure using combinations of Pyramidal GCNs (PGNs) and Long-Short-Term-Memory Units(LSTMs). PGN is a novel way to process non-Euclidean skeletal data in a hierarchical form. Each feature representation in PGN hierarchy is used as an input to an LSTM unit to learn the temporal aspect of the input sequence (shown in Fig.\ref{fig:pipeline} and described in Sec. \ref{sec:ST-PGN}). 

\subsection{Graph Convolutional Network}
Graph convolutional networks (GCN) \cite{zhou2018graph} learn the layer-wise propagation operation that can be applied on structured data represented by a graph. To briefly introduce how GCNs work, assume we have an undirected graph with $N$ nodes, a set of edges between nodes, an adjacency matrix $\mathbf{A} \in \mathbf{R}^{N \times N}$, and a degree matrix $\mathbf{D}_{i i}=\sum_{j} \mathbf{A}_{i j}$. If $\mathbf{x} \in \mathbf{R}^{f\times N}$ represents the feature matrix of the graph ($\mathbf{x}_{i} \in \mathbf{R}^{f}$ is the feature vector of node $i$), a linear formulation of graph convolution is,
\begin{equation}\label{eq:graph1}
    \mathbf{f}=\hat{\mathbf{D}}^{-\frac{1}{2}} \hat{\mathbf{A}} \hat{\mathbf{D}}^{-\frac{1}{2}} {\mathbf{x}_i}^{\top} \mathbf{W},
\end{equation}
where $\hat{\mathbf{A}}=\mathbf{A}+\mathbf{I}$, $\mathbf{I}$ is the identity matrix and $\mathbf{W} \in \mathbf{R}^{f \times c}$ is the weight matrix. So, if the input to a GCN layer is $f\times N$ the output would be $N\times c$. As with any other convolution layer we can have a stack of GCNs each followed by a nonlinear function (such as ReLU) \cite{kipf2016semi}.

In this work, we are following the spatial configuration partitioning introduced in ST-GCN \cite{yan2018spatial}, therefore, $\hat{\mathbf{A}}=\sum_{a}\mathbf{A}_a$ and equation \ref{eq:graph1} is written in a summation form.
\begin{equation}\label{eq:graph2}
    \mathbf{f}=\sum_a\hat{\mathbf{D}}_a^{-\frac{1}{2}}\mathbf{A}_a \hat{\mathbf{D}}_a^{-\frac{1}{2}} {\mathbf{x}}^{\top} \mathbf{W}_a,
\end{equation}
Eq. \ref{eq:graph2} is represented for $k^{th}$ level of the pyramidal hierarchy in line \ref{alg:GCN} of Algorithm \ref{alg:FeatureUpdate}. We hypothesize that a hierarchical graph convolution that operates on human joints, body parts and global structure would enrich the input representation. 

\subsection{Pyramidal Graph Architecture}\label{sec:HGCN} 
Pyramidal Graph Convolutional Network (PGN) is a hierarchical GCN that produces different spatial features with semantic meaning at different levels. The input to the PGN is the skeleton with $\mathbf{N}$ joints represented by a tensor ($\mathbf{X}$) of dimension $F \times N \times T$. Each GCN aggregates features along the spatial dimension using a specific adjacency matrix $\hat{\mathbf{A}}_k$ using Eq.\ref{eq:graph2}. Our PGN has three graph levels ($\hat{\mathbf{A}}_1$,$\hat{\mathbf{A}}_2$,$\hat{\mathbf{A}}_3$). The initial GCN works on the skeleton with $\hat{\mathbf{A}}_1$, which is constructed based on the skeleton connections and accompanied with an edge-importance matrix. The subsequent graph levels represent the body parts and global structure respectively. Since the correlation between the nodes for higher level graphs is unknown, $\hat{\mathbf{A}}_2$ and $\hat{\mathbf{A}}_3$ represent fully connected graphs and we let the edge-importance learn the correlations.

 Thus our model has a hierarchy of graphs with the base as the input skeleton and the top level a graph with three nodes representing right arm and leg, left arm and leg, and the head and spine. 
 We refer to this hierarchical graph structure as a pyramidal graph architecture because it is large at the base and becomes smaller as we move to the top levels.
\subsection{Group Average Pool}\label{sec:GAP}
A Group Average Pool (GAP) layer average-pools the features in a selected group of nodes/joints using a specific kernel ($\mathbf{J}_k$) for each level (line \ref{alg:GAP} of Alg. \ref{alg:FeatureUpdate}). The resulting graph has nodes that represent a higher level body part (as shown in Fig. \ref{fig:pipeline}). Therefore, every layer of the pyramid has a semantic meaning, from low to high level. In the bottom left corner of Fig. \ref{fig:pipeline}, we show how the groups are defined in TUM and UW-IOM datasets.  

More specifically, feature masking is inspired by \cite{dai2015convolutional} which is generally used in foreground background separation. Here, kernels are pre-determined matrices with ones or zeros. These kernels are element-wise multiplied by the features to group only certain body parts one at a time.  For example, the kernel has ones in the particular rows corresponding to those joints representing left arm ($7,9,11$) and zero everywhere else. Hence, the masked features (features multiplied by the mask) all belong to the left arm. These features are average pooled as they belong to the same group.  Multiple such combinations are used to group the joints into different parts. Similarly , parts are combined into global structure using another set of kernels. Such successive GCN-GAP combinations allows us to model the entire local and global motions jointly. We refer to this as the feature update rule (Alg.\ref{alg:FeatureUpdate}), and later in Sec. \ref{sec:ST-PGN}, it is referred to as a bottom-up pathway.

\begin{algorithm}[h]
\caption{Feature Update Rule}\label{alg:FeatureUpdate}
\begin{algorithmic}[1]\footnotesize
\State $\mathbf{X}_0 \gets \mathbf{X}$\Comment{input skeleton distributed over time}
\State $k \gets 1$\Comment{iterator}
\While {$k \leq3$}
\State $\mathbf{f}_k=\bm{c_k}(\mathbf{X}_{k-1};\hat{\mathbf{A}}_k)$\label{alg:GCN} \Comment{GCN operation}
\State $\mathbf{X}_k=\begin{cases}
\bm{g_k} (\mathbf{f}_k;\mathbf{J}_k) & \text{if $k<3$,}\\
None & \text{otherwise.} 
\end{cases}$\label{alg:GAP} \Comment{GAP operation}
\State k=k+1
\EndWhile\newline
\Comment{The $\mathbf{f}_k \forall k \in (1,2,3)$  are used as input features for the feature pyramid operations in Algorithm \ref{alg:pyramidUpdate}.}
\end{algorithmic}
\end{algorithm}
\vspace{-2em}
\begin{algorithm}[h]
\caption{Pyramid Update Rule}\label{alg:pyramidUpdate}
\begin{algorithmic}[1]\footnotesize
\State $k \gets 1$\Comment{iterator}
\While {$k \leq3$}
\State $\mathbf{p}_k=$$\begin{cases}
\bm{w_k} \otimes \mathbf{f}_k & \text{if $k<3$,}\\
\mathbf{f}_k & \text{otherwise.} 
\end{cases}$ \Comment{$1 \times 1$ convolution}
\State $\bm{z}_k=\begin{cases}
        \bm{p}_k \oplus \bm{u}_k(\bm{p}_{k-1}) & \text{if $k<3$,}\\
        \bm{p}_k & \text{otherwise.} 
\end{cases}$
\State \Comment{\text{Upsample \& Add}}
\State k=k+1
\EndWhile\newline
\Comment{The Following $\bm{z}_k \forall k \in (1,2,3)$  are used as input features for the temporal modelling using three separate LSTMs.}
\end{algorithmic}
\end{algorithm}
\vspace{-2em}
\begin{table}[h]
\centering
\footnotesize
\begin{tabular}[t]{ll}
\hline
Symbol & Legend\\
\hline 
$N$ &  Number of 3D Skeleton joints,$(x,y,z)$ tuples \\
$T$ &  Time history of 80 samples  \\
$\mathbf{c}_k$ & Graph convolution(GCN) at each hierarchy k \\
$\hat{\mathbf{A}}_k$ & Adjacency matrix at each hierarchy k\\
$\mathbf{X}_0$ & Input Skeleton feature to first GCN ( $3 \times N \times T$)\\
$\mathbf{g}_k$ & Group Average Pool at each hierarchy k\\
$\mathbf{J}_k$ & Pooling kernel at each hierarchy k\\
$\mathbf{f}_k$ & Output of each GCN  ( $3 \times N \times F$) \\
$\mathbf{w}_k$ &$1 \times 1$ convolution operation\\
$\mathbf{u}_k$ & Upsample and Add\\
$\mathbf{p}_k$ & Output of $1 \times 1$ convolution\\
$\mathbf{z}_k$ & Final features sent to LSTMs\\
\hline
\end{tabular}
\caption{Description of the symbols used in Algorithms}
\end{table}%
\subsection{Feature Pyramid Graph Convolutional Network} \label{sec:ST-PGN}
Feature pyramids have been an important component of object recognition algorithms \cite{lin2017feature,shrivastava2016training, he2016deep}. The advantage of using pyramids is that it produces a multi-scale feature representation in which all feature levels are semantically strong. Especially in skeleton-based action recognition the correlation of body-parts can be very informative in recognizing actions. However, a pre-defined graph might not be sufficient to represent every sample. For example, in ST-GCN graph, there is no connection between hand and head, which is important in actions such as eating. Therefore, here we are generalizing the feature pyramid network to a GCN pyramidal feature hierarchy, and we believe that learning the correlations at different levels of the hierarchy enhances the performance of our model. Here feature pyramids are still valid in skeleton structure as global motion is a combination of local motion of parts and part motion is a combination of local motion of joints. Hence our feature pyramids aggregate joints, parts and global features jointly \cite{xu2018unsupervised}.

The feature pyramid networks consist of two pathways, a bottom-up and a top-down pathway.  The \textbf{bottom-up pathway} is the feed-forward computation of the backbone GCN, which computes a feature hierarchy consisting of feature maps at different scales. The \textbf{top-down pathway} produces higher resolution features by up-sampling spatially larger, but semantically stronger, feature maps from higher pyramid levels. The top-down path is enhanced by the features produced in the bottom-up pathway through lateral connections. The features from the bottom-up pathway undergo a $1\times 1$ conv layer to reduce channel dimensions and then are merged into the top-down pathway features by element-wise addition. The purple connections in Fig. \ref{fig:pipeline} shows this process, and it is described as the pyramid update rule.
\subsection{Spatio-Temporal Modelling}
\begin{figure*}[t]
\centering
\includegraphics[width=0.9\linewidth]{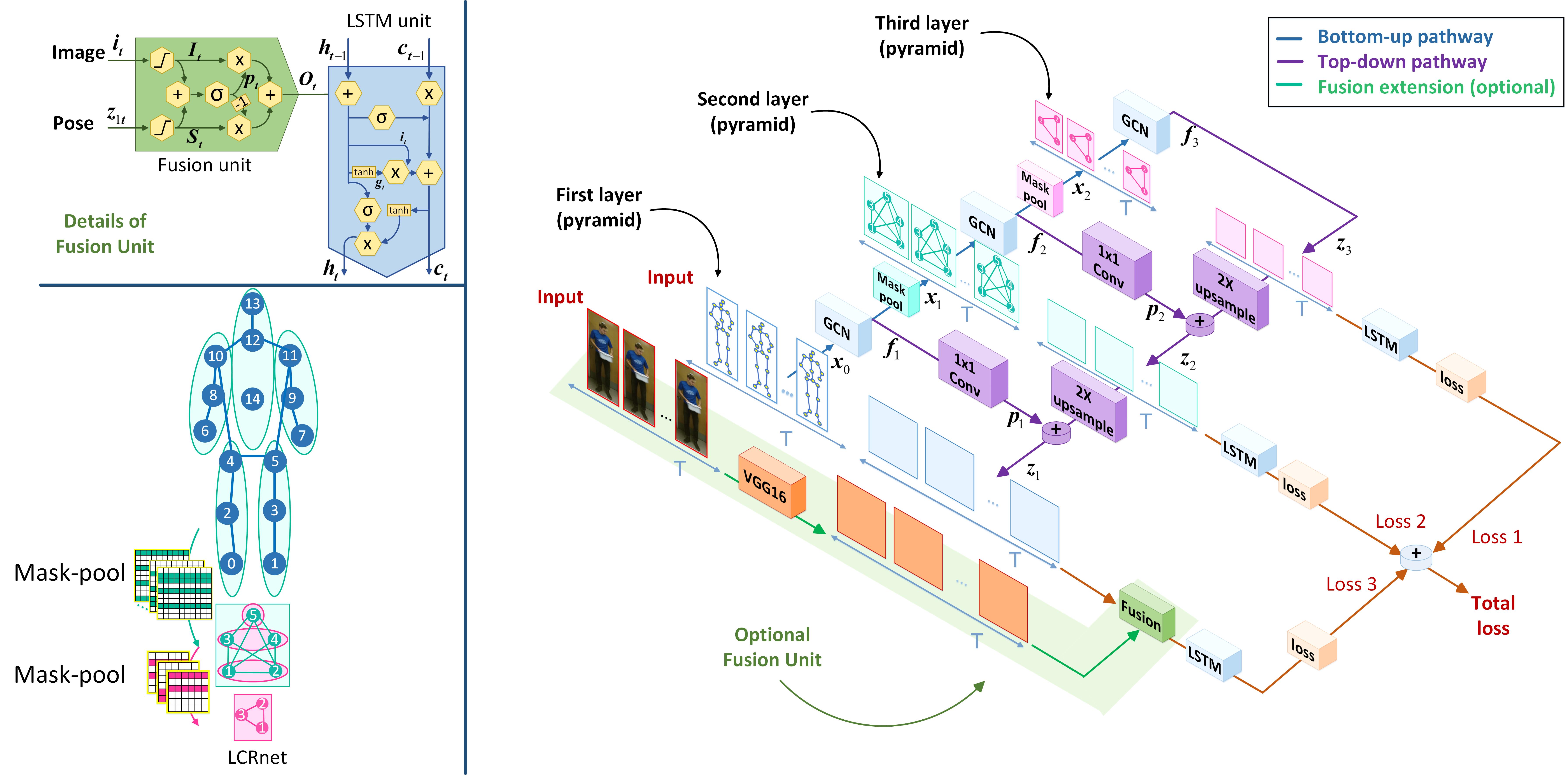}
\caption{The Feature Pyramid Convolutional Graph Network pipeline.}
\label{fig:pipeline}
\end{figure*}

Now we briefly summarize ST-PGN steps that are described in Algorithm \ref{alg:FeatureUpdate}-\ref{alg:pyramidUpdate} and Fig. \ref{fig:pipeline}, and also describe major differences with respect to ST-GCN. The input skeleton ($\mathbf{X}_0$) goes through three levels of GCN and GAP, and the output of each level ($\mathbf{f}_k$) is aggregated with the upsample features through lateral connection and forms the final features ($\mathbf{z}_k$). Each pyramidal feature is passed through separate LSTMs to create three frame-wise activity predictions. As an ablation study we either 1) average these  three predictions and compute one loss or 2) compute three losses separately and average the predictions while testing. The latter gives us better performance.
As a comparison, in ST-GCN, the input goes through a sequence of multiple GCN and TCN units so that the final feature embodies spatial and temporal properties of the input. A final feature that summarizes spatial and temporal properties is the key for video clip classification. However, when we need to recognize activities frame-wise, that strategy fails as will be shown in Sec. \ref{sec:preformace}. Therefore, we are extracting the spatial features through PGN and send these features to individual LSTM units so that the temporal aspect is learned at different spatially semantic layers.
\subsection{Ergonomic Risk Assessment}
Given the input skeleton ($\mathbf{X}$)  and the recognized activity, we compute REBA. \cite{parsa2019toward} averaged the score over all subjects offline and reported one score for each activity class. However we compute the score online and adjust it using the model prediction as shown in Fig. \ref{fig:test_pipline}. For additional details on REBA computation, refer to \cite{hignett2004rapid}.
\subsection{Optional Fusion Unit}
To study the benefit of image features, we also perform experiments with image features concatenated along with skeleton pose features.  We hope to avoid confusion in situations with object handling. Hence we extract VGG16 features from a crop image region around the human and fuse them with the final skeleton feature pyramid. Our fusion unit is inspired by GRU~\cite{dey2017gate}, that learns to weight the features before LSTM. We freeze the weights of the pre-tained network and only train the fusion unit along with the final LSTM layer. While the benefit of the image features are very minimal, for completeness, we will describe the fusion unit below.

At time $t$, let the image features and the final feature pyramid layer features be denoted by $i_t$ and  ${z_1}_t$, respectively.  Since the dimensions of these features do not match, we apply linear weights ($\bm{U_i}$, $\bm{U_z}$) to transform them into the same dimension and arrive at the transformed image and skeleton features $I_t$ and $S(t)$ as shown in Eq.~\ref{IandS}. The terms $W_i$ and $W_z$ are learnt weights that are used to learn a gauging value ($p_t$) between the two features similar to the GRU. The weight $p_t$ is squished to take on values $\in [0,1] $ using a sigmoid operation. Finally this weights are multiplied to the incoming features. 
\begin{align}
    I_t&=\text{relu}(\mathbf{U}_i \ast i_t) ,S_t=\text{relu}(\mathbf{U}_z \ast {z_1}_t) \label{IandS}\\
p_t &= \sigma(\mathbf{W}_i \ast I_t + \mathbf{W}_s \ast S_t),\label{eqn:fusion-gating} \\
\mathbf{O}_t &= p_t  I_t + (1-p_t) S_t
\label{eqn:fusion-updates}
\end{align}
Where, $\mathbf{O}_t$ is the \textbf{weighted} feature that is sent as input into one LSTM unit. For Example, If $p_t$ is $0.6$ then the image features ($I_t$) is weighted higher and the skeletal features ($S_t$) are weighted lower ($1-0.6=0.4$).
\section{Experiments}
\begin{table*}[]
\centering
\resizebox{\textwidth}{!}{%
\begin{tabular}{|l|l|l|l|l|l|l|l|}
\hline
\multirow{2}{*}{\textbf{Modalities}} & \multirow{2}{*}{\textbf{Backbones}} & \multicolumn{3}{c|}{\textbf{UW-IOM}} & \multicolumn{3}{c|}{\textbf{TUM}} \\ \cline{3-8} 
 &  & mAP (\%) & Edit (\%) & F1-overlap (\%) & mAP (\%) & Edit (\%) & F1-overlap (\%) \\ \hline
\multirow{10}{*}{Skeleton (only)} 
 & \textit{Frame based} & 39.82 $\pm$ 1.45 & 29.26 $\pm$ 1.32 & 37.87 $\pm$ 1.82 & 29.79 $\pm$ 4.74 & 27.55 $\pm$ 2.89 & 32.63 $\pm$ 4.66 \\ \cline{2-8} 
 & \textit{LSTM} \cite{hochreiter1997long} & 79.35 $\pm$ 4.55 & 77.82 $\pm$ 6.34 & 85.32 $\pm$ 5.37 & 44.24 $\pm$ 5.97 & 56.46 $\pm$ 5.92 & 57.13 $\pm$ 8.24 \\ \cline{2-8} 
 & \textit{TCN} \cite{bai2018empirical} & 57.72 $\pm$ 6.40 & 56.40 $\pm$ 5.36 & 64.78 $\pm$ 6.38 & 30.61 $\pm$ 5.40 & 51.07 $\pm$ 6.17 & 49.87 $\pm$ 11.01 \\ \cline{2-8} 
 & \textit{ED-TCN} \cite{lea2017temporal} & 60.05 $\pm$ 4.89 & 81.73 $\pm$ 2.44 & 84.60 $\pm$ 2.64 & 28.89 $\pm$ 5.77 & \textbf{56.75 $\pm$ 8.50} & 55.92 $\pm$ 11.11 \\ \cline{2-8} 
 & \textit{ST-GCN} \cite{yan2018spatial} & 66.94 $\pm$ 3.49 & 61.89 $\pm$ 3.56 & 71.08 $\pm$ 2.83 & 34.73 $\pm$ 5.98 & 53.88 $\pm$ 5.53 & 53.52 $\pm$ 7.09 \\ \cline{2-8} 
 & \textit{ST-GCN+IMP} \cite{yan2018spatial} & 73.28 $\pm$ 4.30 & 67.21 $\pm$ 6.05 & 76.58 $\pm$ 4.95 & 34.93 $\pm$ 4.75 & 52.27 $\pm$ 3.99 & 52.60 $\pm$ 5.72 \\ \cline{2-8} 
 & \textit{GCN+LSTM+IMP} & 81.97 $\pm$ 7.34 & 72.25 $\pm$ 7.24 & 82.04 $\pm$ 6.08 & 45.92 $\pm$ 4.19 & 52.07 $\pm$ 4.01 & 55.26 $\pm$ 5.54 \\ \cline{2-8} 
 & \textit{\textbf{ST-PGN+LSTM (ours)}} & 86.33 $\pm$ 2.71 & 77.92 $\pm$ 2.44 & 86.83 $\pm$ 1.74 & 48.02 $\pm$ 4.68 & 55.31 $\pm$ 5.09 & 57.58 $\pm$ 6.38 \\ \cline{2-8} 
 & \textit{\textbf{ST-PGN+LSTM+IMP (ours)}} & 85.92 $\pm$ 1.62 & 77.75 $\pm$ 2.46 & 86.21 $\pm$ 1.91 & 42.74 $\pm$ 1.03 & 47.19 $\pm$ 6.39 & 51.14 $\pm$ 6.94 \\ \cline{2-8} 
 & \textit{\textbf{ST-PGN+LSTM+IMP+ML (ours)}} & \textbf{87.03 $\pm$ 2.85} & \textbf{97.86 $\pm$ 2.15} & \textbf{87.95 $\pm$ 1.54} & \textbf{49.62 $\pm$ 6.10} & 56.10 $\pm$ 4.98 & \textbf{57.60 $\pm$ 6.03} \\  \hline\hline
\multirow{3}{*}{Image (only)} 
& \textit{Frame based} & 51.62 $\pm$ 4.12  & 25.60 $\pm$ 1.55  & 34.17 $\pm$ 3.08  & 35.33  $\pm$ 5.26 & 28.33  $\pm$  1.94 & 35.34  $\pm$  2.65  \\ \cline{2-8} 
& \textit{LSTM} & 66.50 $\pm$ 7.55  & 48.31 $\pm$ 5.90 & 57.81 $\pm$ 6.64 & 49.04  $\pm$ 7.03 & 52.64  $\pm$  7.50 &\textbf{ 58.60  $\pm$  7.53}  \\ \hline
\multirow{3}{*}{Fusion} 
& \textit{Frame based+ Concat} & 50.54 $\pm$ 1.55 & 27.57 $\pm$ 0.96 & 36.42 $\pm$ 2.09 &  41.70  $\pm$ 5.76 & 29.66  $\pm$  1.25 & 36.04  $\pm$  1.59
 \\ \cline{2-8} 
& \textit{LSTM+ Concat} & 83.55 $\pm$ 5.74 & 72.98 $\pm$ 7.32 & 77.89 $\pm$ 11.70  &  48.71  $\pm$ 9.42 & \textbf{54.86  $\pm$  6.83} & 57.11  $\pm$  8.81
  \\   \cline{2-8}
& \textit{\textbf{ST-PGN+LSTM+IMP+ML+GRU-Fusion (ours)}} & \textbf{87.05 $\pm$ 3.47} & 8\textbf{0.90 $\pm$ 2.06} & \textbf{88.08 $\pm$ 1.89}  & \textbf{ 57.79  $\pm$ 6.43} & 54.49  $\pm$  5.59 & 58.35  $\pm$  9.78  \\ \hline
\end{tabular}%
}
\caption{mAP, edit, and F1-overlap score represented in mean and standard deviation over five splits in UW-IOM and TUM datasets for different methods and modalities. The best results in skeleton and fusion modality are shown in bold.}
\label{tab:results}
\end{table*}

\subsection{Datasets}
The skeletal information is required to construct the graph structure and node features. For our vision only system, we use state of the art 3D skeleton estimation LCR-Net \cite{rogez2017lcr} to estimate poses for the TUM Kitchen and UW-IOM dataset. While the focus of our work is to evaluate our proposed method on online ergonomic datasets,  we also run experiments on an offline setting for Skeleton Kinetics and NTU-RGB datasets by substituting ST-GCN with our network. These experiments and results are provided in the Appendix.\\  
\textbf{UW-IOM Dataset} 
UW-IOM is a new dataset introduced in \cite{parsa2019toward} with the intention of capturing activities that are common in warehouses; therefore, videos consist of three times repetition of a sequence of object manipulation. This dataset has twenty videos recorded using a Kinect Sensor for at an average rate of twelve frames per second. The duration of every video is approximately three minutes. The labels are of four-tier hierarchy, the first tier indicates the object (box/rod), the second tier denotes human motion (walk, stand, and bend), the third tier captures the type of object manipulation if applicable (reach, pick-up, place, and hold), and the fourth tier represents the relative height of the surface where manipulation is taking place (low, medium, and high).\\
\textbf{TUM Kitchen Dataset} The TUM Kitchen dataset \cite{tenorth2009tum} consists of nineteen videos of a sequence of kitchen activities. Four different monocular cameras recorded the activity of an individual with the rate of twenty-five frames per second and the average duration of the videos is about two minutes. Some of the activities we see in these videos are walking, picking up, and placing utensils to and from cabinets, drawers, and tables. We use the provided two-tier labels for this dataset by \cite{parsa2019toward}, which includes a motion verb (place, reach, stand), and a location (cabinet, drawer) or object manipulation mode (both-hands, one-hand). Using these labels, we have twenty-one activity classes. For our experiments, we choose camera two view alone.
\subsection{Implementation Details}
In our experiments, we sample a fixed length $T$=80 frames from each skeleton sequence as the input for online experiments. For offline experiments( NTU dataset and Skeleton Kinetics ) we set the length $T$ = 150 to cover the entire sequence for one label. We set the batch size to $128$ and $32$ for online and offline experiments respectively. In order to compare fairly with ST-GCN, the graph partitioning for the first adjacency matrix ($\hat{\mathbf{A}}_1$) is set to the same spatial strategy and partitioned into 3 subsets: the root node itself, centripetal group, and centrifugal group. However for the subsequent graphs $\hat{\mathbf{A}}_2$ and $\hat{\mathbf{A}}_3$ we assume that fully connected graph as initialization (all nodes are connected to every other node) and learn the edge importance weighting.

It should be noted that we do not modify the original ST-GCN model in terms of number of GCN or parameters. Our final model has only three GCN layers as opposed to the ten GCN-TCN components. More specifically the first GCN layer has $64$ channels, second GCN has $128$ and third has $256$ channels. During training, we use the Adam optimizer [11] to
optimize the network. We set the betas to 0.9 and 0.999 and set weight decay to zero. We split the training and validation using a \textbf{five} fold split in both TUM and UW-IOM. We report he mean and variance of all the splits in the results Table~\ref{tab:results}. We also do a grid search for learning rate(lr) from 0.1 to 0.001. On an average, lr of 0.05 performs best on all the splits in both datasets.
\subsection{Results and Discussion}
\subsubsection{Baseline Models}
\textbf{GCN vs Non-GCN Methods}. To see the benefit of temporal analysis we perform experiments that only take skeleton (joint position) or image as input. We use these feature as inputs of a \textit{TCN} \cite{bai2018empirical}, \textit{ED-TCN }\cite{lea2017temporal} and \textit{LSTM} \cite{hochreiter1997long} model. Baselines are trained in an online fashion. We also perform frame based experiments to determine the efficacy of temporal modelling.  It must be noted that no additional convolution or linear layers are used to transform the pose inputs.  \\  
\textbf{ST-GCN variants}.  We showcase the original ST-GCN implementations modified to support online setting by removing the final average pooling layer. Most ST-GCN variants used for spatio-temporal modelling, support recursive GCN-TCN models that pool messages across the overall graph of full skeleton. We also replace the 1x1 TCN convolutions with LSTMs. We refer to this model as\textit{ GCN+TCN}. Edge Importance, as in the original work, is also trained and  is showcased as \textit{ST-GCN+IMP}. LSTMs generally outperforms the TCNs to capture short transition changes in online fashion. Hence for the following experiments we choose to use LSTM as a primary temporal modelling source. \\
\textbf{ST-PGN variants}. Our models are showcased as pyramid-GCN (PGN) models. Similar to the previous section, we choose to train the edge importance for each of the sub graphs. The \textit{predictions are averaged} for \textit{ST-PGN+LSTM} and \textit{ST-PGN+LSTM+IMP} and used to compute a single loss. Alternatively, our final multi-loss (ML) model has three losses, one for each of the pyramids. These losses are averaged and propagated during training. During testing the model's predictions are averaged and used for evaluation. The results for this model is shown in Table.~\ref{tab:results} under \textit{ST-PGN+LSTM+IMP+ML}.  \\
\textbf{Fusion Models}. To evaluate the impact of adding contextual features, we use a fusion mechanism that learns the importance of each feature modalities through a gauging mechanism ($p_t$ in top-left of Fig. \ref{fig:pipeline}).
\vspace*{-4mm}
\subsubsection{Performance Analysis}\label{sec:preformace}
\begin{figure}[t]
\begin{center}
\includegraphics[width=0.8\linewidth]{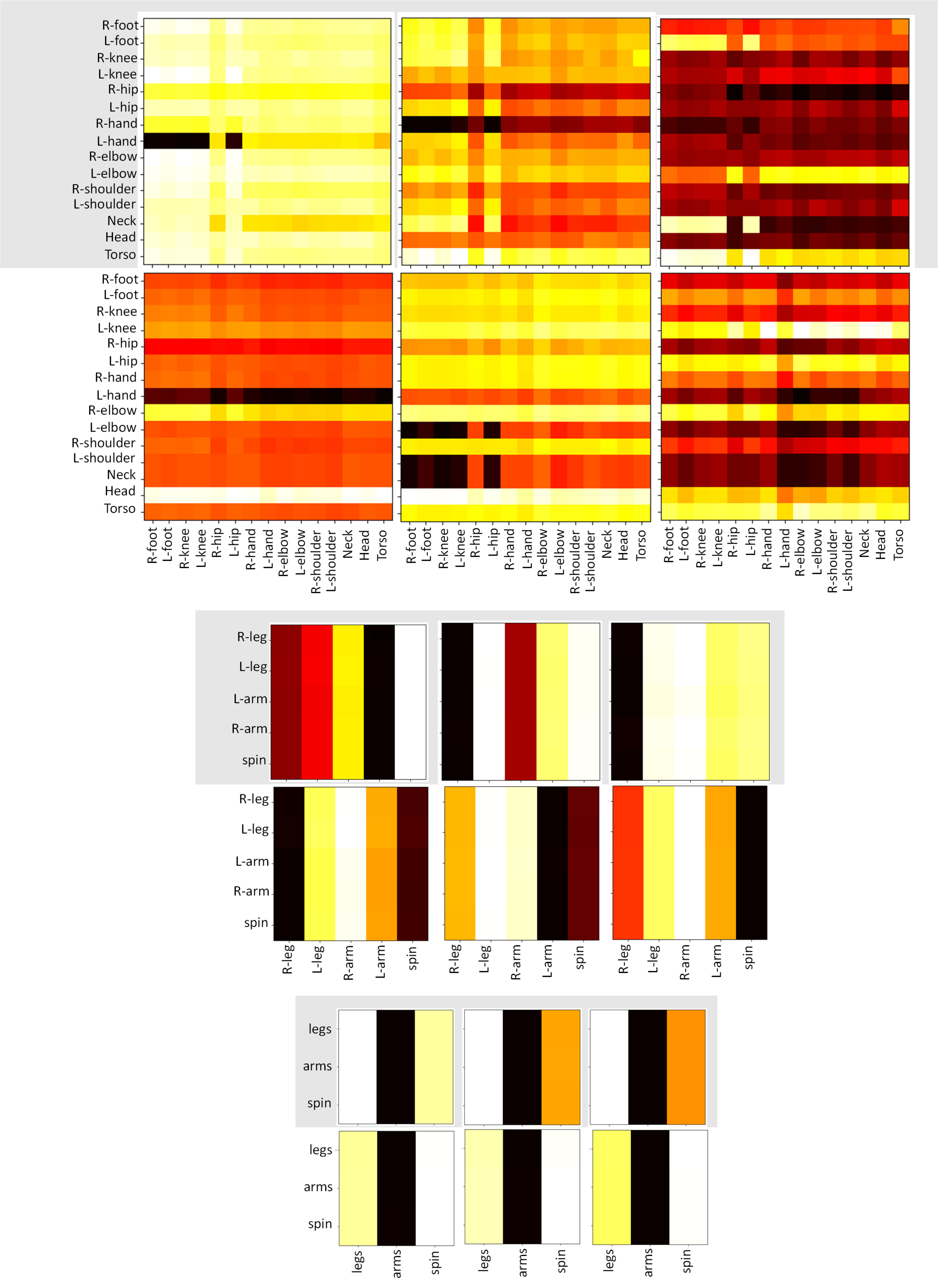}
\end{center}
   \caption{Three level edge importance heat-map in UW-IOM (shaded) and TUM datasets. Each row shows the edge importance of each level of graph pyramid and it is consistent with bottom-left of Fig. \ref{fig:pipeline}. Every level of PGCN consists of the sum of three edge importance multiplied by the adjacency matrix and node features. Here we are depicting the learned edge importance matrices.}
\label{fig:edge_importance}
\end{figure}

The UW-IOM dataset focus is on object manipulation tasks that involve picking up, placing, and carrying objects, as well as walking bending and standing. Therefore, when we look at the edge importance demonstrations in the top left of Fig. \ref{fig:edge_importance}, we see that left hand (L-hand), right hand (R-hand) and right hip (R-hip) are the most important nodes in the low-level edge importance heat-map. Also, at the high-level, the importance of arms is higher than the legs and spine. We achieve an overall +5\% improvement in mAP, +2\% improvement in F1-overlap (\textit{ST-PGN+LSTM}) over the best baseline (\textit{GCN+LSTM} and \textit{LSTM}). However, we see an overall performance boost of +16\% in Edit score and similar to our multi-loss model.  Importantly, ST-PGN is more powerful in distinguishing pick-up and place.  These activities are spatially very similar and differ primarily in temporal aspects.  We do not see a huge benefit in Edit score using our image fusion. However, we see a minor improvement of 1\% in the mAP and F1-overlap.

TUM kitchen dataset also includes object manipulation activities; however, it is focused on common daily activities in a kitchen. Looking at the low-level heat-map (top right in Fig.\ref{fig:edge_importance}) we observe that the hand, elbow, shoulder, and the neck joints have more importance. Looking at the high-level demonstration, we observe that the arms are more important than the legs and spine. We observe an overall improvement in mAP and F1-overlap using our models. However, a simple ED-TCN is slightly better at capturing the sequence and hence the Edit score is higher. Since the subjects move around in the scene, the significance of the lower body(\textit{legs, hip}) is visibly higher in the edge importance compared to UW-IOM.

The results reported in Tab. \ref{tab:results} show that using skeleton is sufficient to get equal or better performance as compared to image-only or the fusion of skeleton and image. In UW-IOM, the human is facing the camera; thus, the detected skeleton is accurate. However, since this is not the case in the TUM dataset, the image-based models perform better as compared to the skeleton only on TUM dataset. If the skeleton is accurate, the addition of the image does not seem to enhance the results significantly in these tasks. 

\begin{figure}
    \centering
    \begin{subfigure}[b]{0.48\linewidth}        
        \centering
        \includegraphics[width=\linewidth]{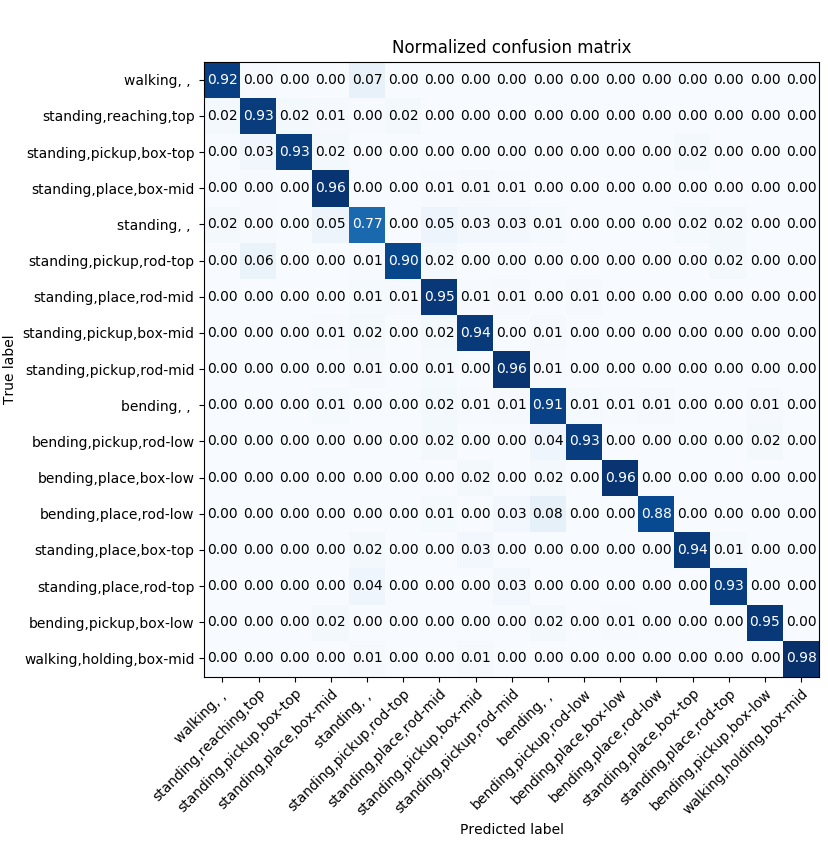}
        \caption{UW-IOM}
        \label{fig:A}
    \end{subfigure}
    \begin{subfigure}[b]{0.48\linewidth}        
        \centering
        \includegraphics[width=\linewidth]{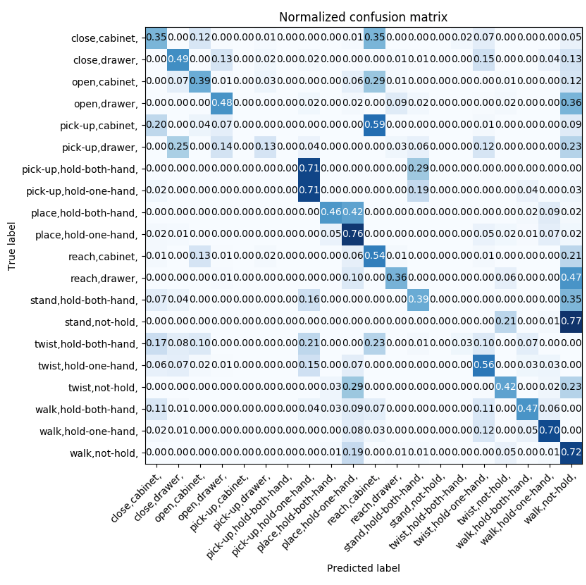}
        \caption{TUM}
        \label{fig:B}
    \end{subfigure}
    \caption{Confusion Matrix of \textit{ST-PGN+LSTM+IMP+ML} model. Larger figures are added in the Appendix section.}
    \label{fig:CMs}
\end{figure}
\vspace{-1em}
\subsubsection{Failure Cases}
We showcase confusion matrices of our best \textit{ST-PGN+LSTM+IMP+ML} model, as described in Figure \ref{fig:CMs} of the previous section. While we see an overall performance increase on both UW-IOM and TUM datasets, the model cannot deal with confusion among similar classes. We showcase the skeleton only model as adding image features do not help significantly. We describe our insights in detail below.\\
\textbf{UW-IOM Dataset}  Our models can differentiate between \textit{box-handling} actions and \textit{rod-handling} actions without the use of image features (The skeleton configuration differs and handling of these objects is distinct due to the object size and location). However, \textit{ Standing} and \textit{walking} misclassifications occur especially when the subject's back faces the camera. Hence important hand motions that help to infer these actions are missed. Better self-occlusion handling is warranted.  Confusion also occurs between \textit{bending} actions such as \textit{bending-place}. This is predominantly due to misclassifications in transitions between these actions since bending is followed by pickup action or preceded by place action. Since it is a challenge for human annotators to accurately label transitions, the edit score should avoid penalizing such transitions.  \\
\textbf{TUM Dataset}  The camera view angle contributes to significant confusion between related classes. We choose the training-validation split with the lowest mAP score to analyze the results.   The following observations are made: \\
1) The \textit{pickup-drawer} and \textit{close-drawer} are completely misclassified in this split. Once the drawer is closed, the pose estimation predicted the hand orientation and location using LCR-Net occlusion strategy \cite{rogez2017lcr}. However, the predicted pose is not always reliable, resulting in poor performance due to incorrect pose input during training. 2) \textit{Walk-not holding} is misclassified for the majority of the classes such as \textit{reach-cabinet}, \textit{reach-drawer}, \textit{stand-hold-both-hands}, \textit{stand-not-hold}. This is attributed to unbalanced class distribution, where most of the actions are walking.  In future work, we plan to address biases introduced by data imbalance by introducing sampling strategies. 3)\textit{ Twisting} actions are very challenging to detect using vision only since we only measure poses in Cartesian coordinates. Adding rotation information should help the model detect twisting actions about certain body axes. 4)\textit{ Pickup-hold-both-hands} gets confused with either \textit{ Pickup-hold-\textbf{one}-hand}  or \textit{stand-hold-\textbf{both}-hands}. Confusion is primarily due to one hand either being occluded by the object being handled, or the pose configuration being too similar in pose configuration with \textit{standing}. More key-points in the pose prediction models could help resolve such issues. 
\section{Conclusion and Future Work}
We proposed a novel Spatio-Temporal Pyramid Graph Convolutional Net-work (ST-PGN) for online action recognition. The method integrates the following:  a) basic prior
knowledge about the skeletal structure, b) hierarchical joint relationships and c) data-driven learning framework for online action based ergonomic risk assessment.  
The proposed approach addresses the simultaneous association of time-varying pose with action and objects interaction to enable downstream applications that involve computational modeling and prediction of various human performance metrics for ergonomic assessment.  

Some open issues remain. First, generalization concerning other skeletal joint representations ( Lie~\cite{vemulapalli2014human}, Quaternion~\cite{pavllo2018quaternet} ) and camera viewpoint changes has not been addressed.  Furthermore, different actions could share similar pose configurations, resulting in severe inter-class confusion.  In future work, we hope to address these issues with improved context fusion, long-term temporal modeling, and biomechanically consistent human pose representations~\cite{zhu2010kinematic}.

{\small
\bibliographystyle{ieee}
\bibliography{References}
}

\end{document}